# Similarities between Arabic Dialects: Investigating Geographical Proximity

Abdulkareem Alsudais*, Wafa Alotaibi, Faye Alomary

*College of Computer Engineering and Sciences, Prince Sattam bin Abdulaziz University, Al-Kharj 11942, Saudi Arabia*

*AAlsudais@outlook.com

## Abstract

The automatic classification of Arabic dialects is an ongoing research challenge, which has been explored in recent work that defines dialects based on increasingly limited geographic areas like cities and provinces. This paper focuses on a related yet relatively unexplored topic: the effects of the geographical proximity of cities located in Arab countries on their dialectical similarity. Our work is twofold, reliant on: 1) comparing the textual similarities between dialects using cosine similarity and 2) measuring the geographical distance between locations. We study MADAR and NADI, two established datasets with Arabic dialects from many cities and provinces. Our results indicate that cities located in different countries may in fact have more dialectical similarity than cities within the same country, depending on their geographical proximity. The correlation between dialectical similarity and city proximity suggests that cities that are closer together are more likely to share dialectical attributes, regardless of country borders. This nuance provides the potential for important advancements in Arabic dialect research because it indicates that a more granular approach to dialect classification is essential to understanding how to frame the problem of Arabic dialects identification.

*Keywords:* Arabic natural language processing, Arabic dialects, geolocation, textual similarity

# 1. Introduction

Much recent research in Arabic Natural Language Processing has focused on studying challenges related to the identification, classification, and processing of Arabic dialects. Arabic is one of the most commonly used languages in the world and contains a rich variety of dialects with marked differences. The language has three main types: Classical Arabic, Modern Standard Arabic (MSA), and Dialectal Arabic (DA) [1]. DA varies between countries and it may also vary between areas in the same country [2,3]. Studying these dialects and their variations has been an ongoing challenge for Arabic Natural Language Processing [4–6]. The primary area of focus for research on Arabic dialects has been the development of methods to classify the dialects. Despite this attention devoted to dialect identification, limited work has examined the effects of the geographical proximity of cities located in Arab countries on dialectical similarity. In this paper, we investigate this topic and explore the relationship between geographical proximity (i.e., distance) and dialect similarity. Our findings contribute an effort to reconsider how current research in nuanced Arabic dialect identification is shaped.

One important issue in Arabic dialects research is how scholars define the scope of Arabic dialects. Put differently, how do scientists select their unit for defining a dialect? For this, some researchers have focused on datasets divided by regions [7–9]. For this specification, a region could be located in one or more countries. The ISO 639-3, which provides a classification for languages, uses this scope for regions and divides Arabic dialects into 30 regions, such as “Gulf Arabic” and “North Levantine Arabic” [10]. These regions tend to be large in terms of the geographical area they cover. Alternatively, others have considered countries as their primary geographical unit [11,12]. Recent dialect research has defined dialects based on geographical areas that are smaller in size than regions or countries (i.e., cities and provinces). This has been motivated by the availability of two corpora that contain texts collected from cities and provinces, which allow for the development of new methods to identify city-level dialects.

These two datasets are NADI [5,6], and MADAR [13]. MADAR is a parallel corpus that was constructed by selecting sentences from the Basic Traveling Expression Corpus (BTEC), which was written in English and French, and then translating these sentences into each dialect included in MADAR. As for NADI, it was first announced as a part of a shared task related to Arabic dialect identification using large data collected from Twitter. The task consisted of two sub-tasks for dialect identification: country-level and province-level. For the latter, the goal was to develop models that accurately identified the specific province from which a dialect had issued from a list of 100 provinces. In the second installment of the shared task, the organizers increased the number of sub-tasks to four. The main difference in this second installment was that for each province, tweets were divided between Modern Standard Arabic (MSA) and dialectal Arabic (DA). Then, two tasks required the identification of country and province for tweets written in MSA and two others required the same, but for tweets written in DA. These two datasets and their shared tasks highlighted the complexity and difficulties associated with developing methods that automatically identify the dialect of text. The way in which this problem is studied is evolving as we are learning more about what defines dialects and their boundaries. Motivated by this recent potential for evolution, we sought to investigate whether additional factors might be pertinent to the study of Arabic dialects. More specifically, we wanted to examine the effects of the geographical proximity between locations on the similarity between the dialects used in these locations. Our inquiry is prompted by the hypothesis that two cities that are commonly grouped within one dialect actually have more in common with nearby locations that are in other countries or regions.

In this study, we are utilizing MADAR and the first release of NADI to investigate the effects of geographical proximity on dialectal similarity between cities and provinces. That is, we want to study whether the distance between locations is correlated with their dialectal similarity. Moreover, we are investigating whether it is possible for the dialects of two locations that are not in the same country to have more in common than two locations that are in the same country. Thus, our research also expands the possibilities for investigating the importance of national borders on dialect formation. Our work relies on the use of cosine similarity to determine the similarity between locations. We are also identifying the distance between locations, and using it to examine the correlation between similarity and distance. We are investigating these topics by utilizing the textual data from several locations in Arabic-speaking countries furnished by MADAR and NADI. In summary, this paper makes several key contributions. First, it examines two commonly used datasets and explores the similarities between written language in the cities and provinces that they include. Then, it investigates the dialectal similarities between cities that are located in the same countries and cities that are not. Accordingly, it tests for the presence of cities with statistically significant positive correlation or high similarity even if they are not located in the same country. Most importantly, it studies the impacts of geographical proximity on dialectal similarity between cities and provinces. Our findings contribute to the study of Arabic dialects identification and understanding.

The rest of the paper is organized as follows: Section 3 lists the research questions. Section 3 includes a review of related work. Section 4 describes the research methodology of the current study on the effects of geographical proximity on textual similarity between Arabic dialects. Section 5 contains the results of the study. Sections 6 and 7 present a discussion of our findings and a conclusion.

## 2. Research Objectives

The main research objective of this study is to study the effects of the geographical proximity of cities located in Arab countries on their dialectical similarity. Specifically, we are investigating the following research questions:

RQ1: What are the similarities between the dialects used in various locations?

RQ2: Are there pairs of cities or provinces that are not located in the same country with similarity values that are higher than the total average of similarity between pairs that are located in the same country?

RQ3: Which pairs have statistically significant correlations?

RQ4: Which locations have statistically significant correlations between similarity and the distance they have with other locations?

RQ5: For the two datasets, is there a statistically significant correlation between similarity and distance?

RQ6: When creating subsets of pairs based on the ranges of distances (e.g., under 100 kilometers (km) and between 100 and 200 km, etc.), are there any significant differences between similarity values for subsets?

# 3. Related Work

## *3.1. Classification of Arabic dialects*

Several researchers have developed new datasets and used them to test models that classify dialects. Some have focused on specific Arabic-speaking regions instead of considering all Arabic-speaking countries. For example, Kwaik et. al [14] developed a Levantine dialect corpus that consisted of data from four countries that are located in the Levantine region (i.e., the region including Syria, Jordan, Palestine and Lebanon). The authors adopted a combination of manual and automatic methods to construct their corpus, including the use of data from Twitter as well as blogs written by individuals who were writing in the dialect of a specified region. In another paper, the authors created a dataset that consisted of spoken Arabic dialects, using YouTube as a source for collection [15]. In a third, a new dataset was constructed by considering five regions: Egypt, the Gulf region, Iraq, the Levantine region, and North Africa [7]. The authors used Twitter to collect tweets from the five regions. They also constructed another dataset from the same regions using Twitter and comments from Facebook and online newspapers [16]. In a similar study that employed a slightly different classification of the regions, Sadat et. al. [8] used “Maghrebi” instead of “North Africa” and also added an “other” category that included only Sudan. They developed their corpus using blog posts and web forums, and then employed it to evaluate the performance of several methods for classifying dialects from a list of 18 countries. Another study relied on the collection of a Twitter dataset from four main regions: the Gulf, the Levantine, North Africa, and Egypt [9]. Another dataset focused on a speech corpus from the Gulf, Egypt, and the Levantine as well as MSA [17]. A dataset entitled “QADI” was similarly constructed based on Twitter data [11]. The data also used 18 countries and contained 540k tweets from 2,525 users. Some studies focused on the North African dialects such as Moroccan [18,19], Algerian [20], and Tunisian [21]. Similarly, others focused on developing NLP solutions based on the dialects of specific counties such as Iraq [22] and Saudi Arabia [12]. Some recent works focused on sentiment classification for dialects [23] and machine translation between Arabic dialects [3]. For example, Farhan et. al. [24] proposed a new method to translate texts between MSA, Egyptian, Jordanian, and Saudi dialects. In summary, these studies focus on segmenting Arabic dialects into either regions or countries. Many studies have since used these datasets and others to develop and evaluate classification methods [25–27].

Recently, several large corpora of texts written in dialectal Arabic and organized according to smaller geographical areas (i.e., cities and provinces) have been introduced. In Abdul-Mageed et. al. [28], the author collected a large dataset from a number of Twitter users whose locations were verified. The dataset was employed in order to predict the dialect (city and country) of each user based on their language use. The authors indicated that the presence of tweets written in MSA affected the results since "MSA is shared across different regions." They demonstrated how the performances for their classifiers increased after they removed tweets written in MSA from the training and testing data. In the aforementioned MADAR and NADI, the two datasets have segmented texts into countries first and then city for MADAR and province for NADI. These two datasets have been utilized in the development of solutions that target dialect identification at the city or province level. For example, ADIDA is a web application developed based on MADAR [29]. The application predicts the dialect of an input text from one of the 25 cities in MADAR. The prediction also includes MSA as an option when the classifier identifies the input text as one written in Modern Standard Arabic. Finally, since their introduction, a variety of methods have been tested in order to classify the texts in MADAR and NADI [30–35].

### *3.2 Studying similarities between Arabic dialects*

Limited previous research has focused on studying the differences between Arabic dialects. In one study, the authors considered several corpora that consisted of country-level dialects and compared their commonalities [36]. The authors studied the different dialects' similarity to Modern Standard Arabic and found that "Levantine dialects are in general the closest to MSA, while the North African dialects are the farthest," despite the fact that none of the corpora used in their research included data from the Gulf region. If a corpus consisting of texts from Gulf countries had been used, different findings might have been obtained. Moreover, unlike in this paper, their work did not consider city-level similarity or study geographical proximity as a factor. The similarity between country-level dialects was also studied by Bouamor et. al. [37]. The authors compared the similarities between texts collected from four countries. In the previously-described paper by Kwaik et al. [36], the authors also compared dialects collected from the four countries in their corpus. They concluded that there is "great overlap between the dialects and dispersion of lexical items between categories," while also acknowledging that there was "a little similarity between the two dialects on the lexical level" when classifying only Jordanian and Lebanese texts. Finally, in the MADAR original paper, the authors considered the similarity between cities in the dataset by calculating the overlap coefficient, which was determined based on "the percentage of lexical overlap between the vocabularies for each dialect pair" [13]. They concluded that "When MSA is not included, the average similarity between the dialects is 26.3%." The researchers behind MADAR also found that Jerusalem and Amman had the highest similarity. The two cities are not in the same country, but ranked third in terms of their geographical proximity when compared to all the other pairs in the dataset. The authors also found that Muscat, which is in Oman, a Gulf country, had the highest similarity when compared to MSA. Furthermore, NADI's winning team for the 2020 country identification sub-task indicated that their model "suffers when trying to differentiate between geographically nearby countries" [33], indicating that country demarcations may be less consequential than geographical distance when considering the characteristics of Arabic dialects. These examples suggest that more research into geographical proximity as an indicator for dialect similarity is needed. In this paper, we are building on this previous work by providing an in-depth investigation into the similarities between city-level dialects and their geographical proximity.

## 4. Methodology

### *4.1. Datasets*

Table 1: The datasets used in this study

| Dataset | # of Countries | # of Cities/Provinces | Average entries per location |
|---|---|---|---|
| MADAR | 15 | 25 | 2000 |
| NADI | 21 | 100 | 259.5 |

In this paper, we used MADAR and the first release of NADI to explore the research questions and provide results that are more likely to be generalizable. For the full lists of names for cities and provinces in the two datasets, we refer to the original papers for the two sets. MADAR consists of dialectical textual contents related to travel. Of the several corpora in MADAR, we used one titled "corpus26," which contains texts from 25 cities that are located in 15 different Arabic-speaking countries. Each city in the dataset has 2000 entries, usually about one sentence long. Some of the entries are questions about topics such as directions or foods. Other entries contain statements, requests, or instructions related to travel or tourism. A crucial distinction between this dataset and NADI is that in MADAR, each entry is written in the dialect corresponding to each of 25 cities studied. In other words, the first entry for “Riyadh” is also the first for “Cairo” with the only difference being that the one for “Cairo” is written in the dialect used in Cairo. Thus, the entire dataset was developed based on human-translated data. Table 1 includes summary statistics on the two sets and their locations.

The second dataset we used in this study, NADI, consists of tweets collected from 100 different provinces that belong to 21 countries. NADI uses “provinces” as its primary unit of measure instead of “cities” and one province may contain multiple cities. However, all provinces are located in one and only one country. While NADI includes data classified as “training,” “development,” and “testing,” we only used the data labelled as “training” and “development” in this study. The data was collected using the Twitter API and it relied on retrieving tweets from users who were identified by the researchers as those who mostly tweet in their local dialect. One limitation of NADI compared to MADAR is the small number of entries that are available for each location. While each city has 2,000 entries in MADAR, each province in NADI has an average of only 259.5 number of tweets. Moreover, some of the tweets in NADI contained text written in Modern Standard Arabic (MSA), which may affect the similarity results. Furthermore, an issue of mislabelled tweets was also described by the team that won NADI 2020’s country identification shared task [33]. The authors cited examples in which this issue was limited not solely to MSA, but included some dialects as well. The authors behind NADI also acknowledged that a few of the tweets in the dataset were written in Farsi. Their recognition that some tweets had also been labelled with the wrong dialect, or were even written in a language other than Arabic, allows us to acknowledge a further limitation to the results of our study. In addition, the authors who won the province identification task stated that certain tweets written in MSA were common and present in several dialects [34], indicating the potential for overlap in the data processed. Nevertheless, the dataset includes a total of 25,957 number of tweets that we used to study the research questions explored in this paper.

## *4.2. Preprocessing*

We performed several preprocessing steps on the datasets. The objective of these measures was to remove elements of the text that could negatively affect the quality of the results. First, we removed diacritical marks from the text. This is a common text preprocessing step in Arabic NLP, practiced in order to eliminate extra characters present in the text. To do this, we simply specified a list of these characters and removed them from both datasets. The second preprocessing step was to normalize several letters because we had observed that the data from NADI included many instances where certain letters were written differently based on the writer, probably because NADI consists of data collected from social media posts, which tend to be written informally. For example, all instances of (آ), (أ), and (إ) were changed to (ا) and all instances of (ى) were changed to (ي). We also noticed that some tweets included repeated characters. In colloquial written speech, some letters are frequently repeated in order to stress certain emotions. To normalize how this type of emotive word was written, we removed repeated letters and replaced all instances with one only. For example, the word (واااااااو) was changed to (واو). We also removed several elements from catalogued tweets that were unique to data from Twitter, like mentioned users, URLs, and emoticons. Additionally, all English letters were removed, as well as all digits written in either Arabic or English. We also used a list of Arabic punctuations and another for English in order to remove all of the listed punctuations from the corpora. Finally, we used a collection of Arabic stop words to remove those that appeared in the datasets. The list we used, which contains 243 stop words, is available in the Python library NLTK [38]. While most of the words from the list seem to be written in Modern Standard Arabic, we observed that a few seem to be from specific dialects. Thus, this step may have removed certain tokens from the text that could be from specific dialects, another potential limitation of the study at hand.

### *4.3. Textual similarity detection*

This study relies on measuring the similarities between texts from several regions located in Arab countries. To determine the textual similarity between two locations, we combined all the textual entries for each location in one document. We completed this process for both datasets. For example, all the tweets from "Riyadh" in NADI were added to one large document. Similarly, all 2,000 entries in MADAR tagged as originating from "Riyadh" were combined in one document labelled "MADAR-Riyadh." Following this process, NADI was transformed into 100 documents, each corresponding to a province, and MADAR into 25, each corresponding to a city. Then, we used the Python library Scikit-learn [39] to generate a TF-IDF representation for all the documents in each dataset. We did not change any of the default settings that Scikit-learn has for its TF-IDF model. Following this step, we created one matrix for each dataset containing all the locations in its corpus. Then, we populated the matrices by calculating the cosine similarities between each pair of locations. To calculate the similarity, we used the Scikit-learn implementation of cosine similarity. The values for each pair ranged from “0” when there was no similarity at all, to “1” when comparing the same document to itself.

An important step in this study was to determine whether two locations were correlated, and, if they were, to quantify the statistical significance of the correlation. We used the Pearson correlation coefficient and its implementation in the Python library SciPy [40] to measure the correlation between two cities or two provinces. The test gives correlation scores that range from “-1”, when there is a negative correlation, to “1”, when there is a positive correlation. When there is no correlation, the value is “0”. In this study, we generated correlation values by comparing all the similarity scores for one location to all the similarity scores for another location. Thus, while the cosine similarity value only measures the direct likeness between two locations, their correlation value also considers all of their other similarity values. Consequently, these results enabled us to determine whether two regions shared similar scores with all other locations. We also used the p-values as discovered by the test to find all the cities or provinces that have a statistically significant correlation. It is important to note that the p-values generated were based on a two-tailed test. Thus, we added a check to make sure that two locations were actually positively correlated when they had a statistically significant correlation. This process helped us find cities that are not in the same country, but which nevertheless have statistically significant correlation with one another.

### *4.4. Distance determination*

In this paper, we aim to discover the effects of physical proximity on the textual similarity between locations. This should help us determine whether there is a relationship between textual similarity and geographical distance. To begin our investigation, we first collected the coordinates for all the locations in the two datasets. We completed this by using a tool provided by Google[1] that allowed us to search for names of cities and get their coordinates. Subsequently, we constructed a list that contained the longitude and latitude pairs for each location. To measure the distance between cities, we used the Python library Haversine [41][2]. The library uses the Haversine formula, which is a common method to calculate the distance between two locations. For each pair in both corpora, we used the library to find the distance between the pair in km. To ensure that the distances found using the library are accurate, we manually examined a sample of the distances found and confirm their accuracy using the "measure distance" feature in Google Maps[3].

Using the distance between locations alone may have some limitations. For example, it is possible that two locations that are in two neighbouring countries (i.e., countries that share borders) will have dialects that are more similar than two locations that are not in neighbouring countries even if they are closer as determined by the Haversine formula. Nevertheless, by following this formula and the previous steps, we generated a results table for each dataset. The table included the name of the two-location pair, their similarity as determined by cosine similarity, their correlation as determined by

---

[1] https://developers.google.com/maps/documentation/geocoding/overview
[2] https://github.com/mapado/haversine
[3] https://support.google.com/maps/answer/1628031

Pearson, their p-value scores, and their distance in km as calculated by the Haversine formula. The tables did not include results for identical pairs (a city and its results with itself) and it also included each pair only once (e.g., we did not include both Cairo & Sfax and Sfax & Cairo). These two tables are the primary source for most of the results in Section 5.

## *4.5. Research questions*

Table 2: The datasets used in this study

| Research question | Data | Methods |
|---|---|---|
| RQ1: What are the similarities between the dialects of locations? | The pairs of cities and provinces in MADAR and NADI | Similarity |
| RQ2: Are there pairs of cities or provinces that are not located in the same country that have similarities values that are higher than the total average of similarity between pairs in the same country? | The pairs of cities and provinces | Similarity |
| RQ3: What are the pairs that have statistically significant correlations? | The pairs of cities and provinces | Similarity and correlation |
| RQ4: What are the locations that have statistically significant correlations between the similarity and distance they have with other locations? | Individual cities and provinces | Similarity, correlation, and distances |
| RQ5: For the two datasets, is there a statistically significant correlation between similarity and distance? | Entire MADAR and NADI sets | Similarity, correlation, and distances |
| RQ6: When creating subsets of pairs based on the ranges of distances (i.e. under 100 km and between 100 and 200 km etc.), are there any significant differences between similarity values for subsets? | Several sets of pairs in MADAR and NADI segmented based on distances | Similarity, correlation, and distances |

We used these methods and the two datasets to respond to our research questions. Some of the research questions focused on studying pairs and results for individual locations (RQ1, RQ2, RQ3, and RQ4) while others examined attributes for the entire sets or large subsets (RQ5 and RQ6). All the questions were designed in order to provide new insights regarding the effects of geographical proximity on textual similarity between dialects. The similarity values were used to answer all six questions while distance and correlation results were used for only some of the questions. Table 2 lists the research questions, the data they target, and the methods we used to answer them.

## 5. Results

Table 3: Summary statistics for the two datasets

| Dataset | # of pairs | # of pairs in the same country | #of pairs that are not in the same country |
|---|---|---|---|
| MADAR | 300 | 12 (4%) | 288 (96%) |
| NADI | 4,950 | 410 (8.2%) | 4,540 (91.8%) |

This section explains the results obtained when we investigated the two datasets. First, Table 3 includes a summary of the statistics for the two datasets. There were 300 pairs in MADAR, while NADI included 4950. The table also lists the number and percentage of pairs that are in the same country as well as those that are not in the same country. For the former, the percentages are calculated by dividing the number of pairs with two cities that are in the same country by the total number of pairs. When comparing the two datasets, it appears that NADI has a higher percentage than MADAR of pairs in the same country. This difference may have an effect on the results. For example, average similarities may vary significantly between the two.

## *5.1. Investigating similarity (RQ1)*

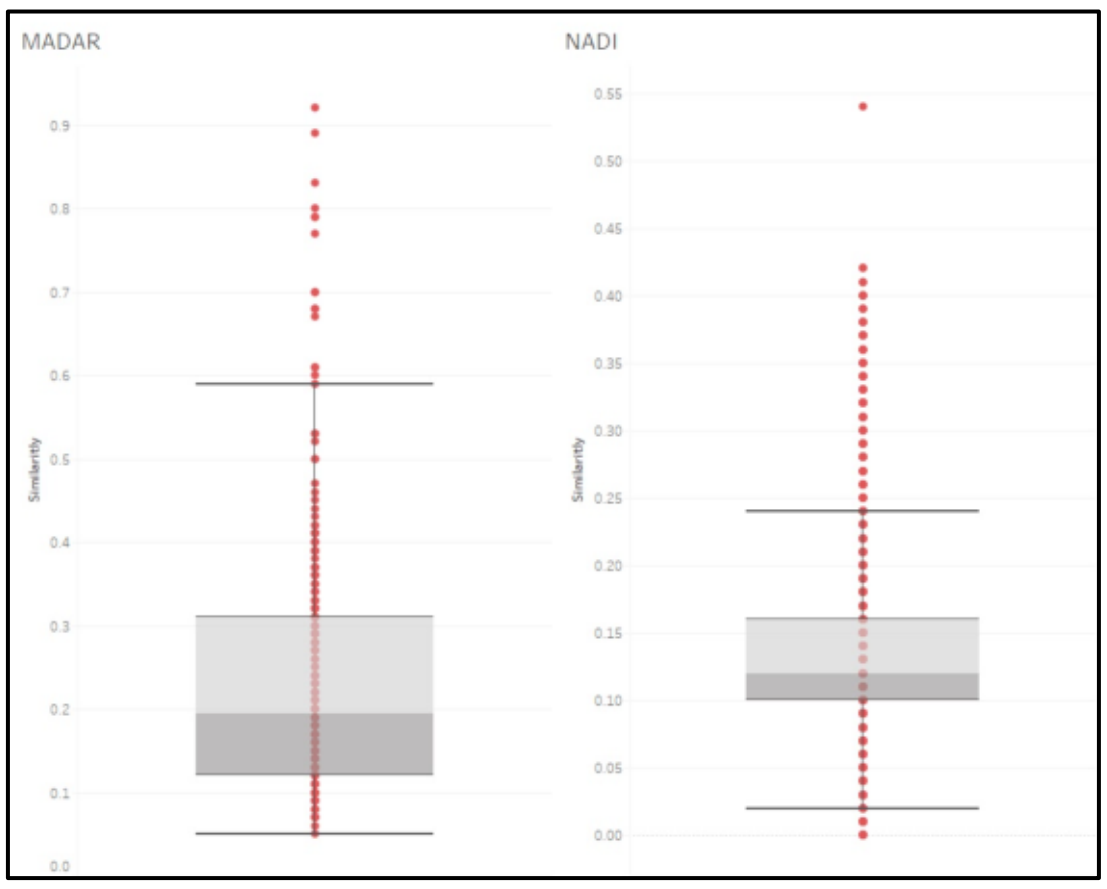

**Figure 1.** Box-and-whisker plots of similarity values between pairs in MADAR and NADI

Table 4: Similarity results

| Dataset | Average similarity | Highest similarity | Lowest similarity |
|---|---|---|---|
| MADAR | 24% | 92% | 0.05% |
| NADI | 13.2% | 54% | 0% |

We first calculated all the similarity scores for pairs in the two datasets, the results of which are in Figure 1 and Table 4. The table includes the highest and lowest pairs in terms of similarity. The results indicated that pairs in MADAR had an average similarity of 24% while the average was 13.2% for pairs in NADI. As described in Section 4.1., entries for cities in MADAR were completed by translating the same original source. Thus, this higher percentage of average similarity is likely due to the recurrence of the same topics within the documents. For example, the Arabic word for "hotel" could be the same for many dialects, and thus its presence in many cities may increase the average similarity. In summary, these results respond to the first research question which is concerned with the overall similarities between pairs.

### *5.2. Similarity between pairs in the same country and pairs not in the same country (RQ2)*

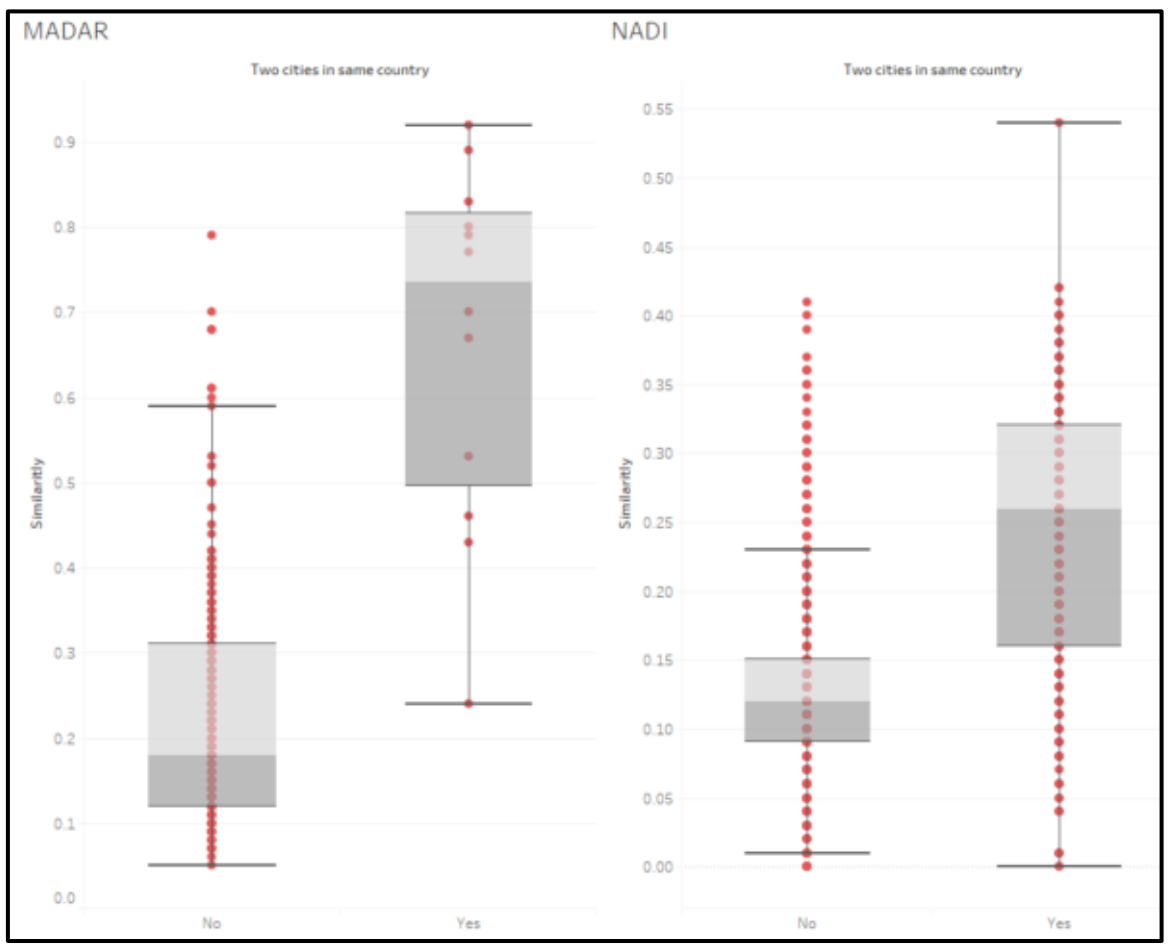

**Figure 2.** Box-and-whisker plots showing the similarity values related to RQ2

The two datasets also varied regarding the average similarity between the group of pairs located in the same country and a second group of pairs not located in the same country. In MADAR, the average similarity for the first group was 66.9%, while it was 22.2% for the second group. Likewise, pairs found in the same country had an average similarity of 24% in NADI and those that were not had an average similarity of only 12.2%. This represents the first, and perhaps expected, finding of this work: Cities that are located in the same country had an average similarity that is higher than the total average similarity for all pairs. Figure 2 displays two box-and-whisker plots of the pairs in the two datasets and results for pairs in the same countries and pairs not in the same countries.

Following our assurance that this was the case, we wanted to determine whether there were pairs with similarity values that were higher than the average similarity for pairs that were in the same country. The existence of such pairs would be an indicator that geographical proximity could be related to textual similarity. To investigate, we processed similarity values for all pairs and identified those that had values that were above 66.9% for MADAR and 24% for NADI. Results indicated that there were four pairs in MADAR that met this condition. The four pairs were: Damascus & Salt, Damascus & Amman, Salt & Jerusalem, and Jerusalem & Amman. As for NADI, there were a total of 60 pairs. However, most of these pairs were included in three provinces. The three were: "South Lebanon, Lebanon," "Tanger Tetouan, Morocco," and "Oran, Algeria." When we removed pairs that included any of these three, the lists were reduced from 60 to only 10.

## *5.3. Pairs with statistically significant correlation (RQ3)*

Table 5: Correlation and Similarity results

| Dataset | Average correlation | Pairs with SS Positive correlation | Pairs with SS Positive correlation (same countries) | Pairs with SS Positive correlation (Different countries) |
|---|---|---|---|---|
| MADAR | 0.05 | 43 (14.3%) | 11 (out of 12) | 32 |
| NADI | 0.08 | 811 (16.3%) | 298 (out of 410) | 513 |

To identify cities with statistically significant positive correlation, we used two similarity matrices. These matrices included the cities in the header and first column. Each row included the similarity values for the cities with all other cities. Then, we compared each pair of rows. Ultimately, we provide a similarity value calculated by comparing text, and a correlation value calculated using the two lists that included all of their similarity values with other cities. After we specified 0.05 as the significance level, we found 43 of the 300 pairs in MADAR having a statistically significant positive correlation. Eleven of these pairs were for cities located in the same country, which indicated that all but one of such pairs from MADAR had a statistically significant positive correlation. When we decreased the significance level to 0.01, the number decreased from 43 to 29 and all dropped pairs were from different countries. As for NADI, there were 811 pairs with SS positive correlation at 0.05 SL. We noticed that NADI had some pairs with SS negative correlation, which we did not include in the 811. The table shows how many of these 811 were from the same country. In total, 513 of the 811 pairs were for provinces that were not in the same country. The main finding from these results is that there were many pairs from different countries that still had statistically significant positive correlation (i.e., they were comparable in terms of their similarity values with other cities).

## *5.4. Investigating similarity and distance for cities individually (RQ4)*

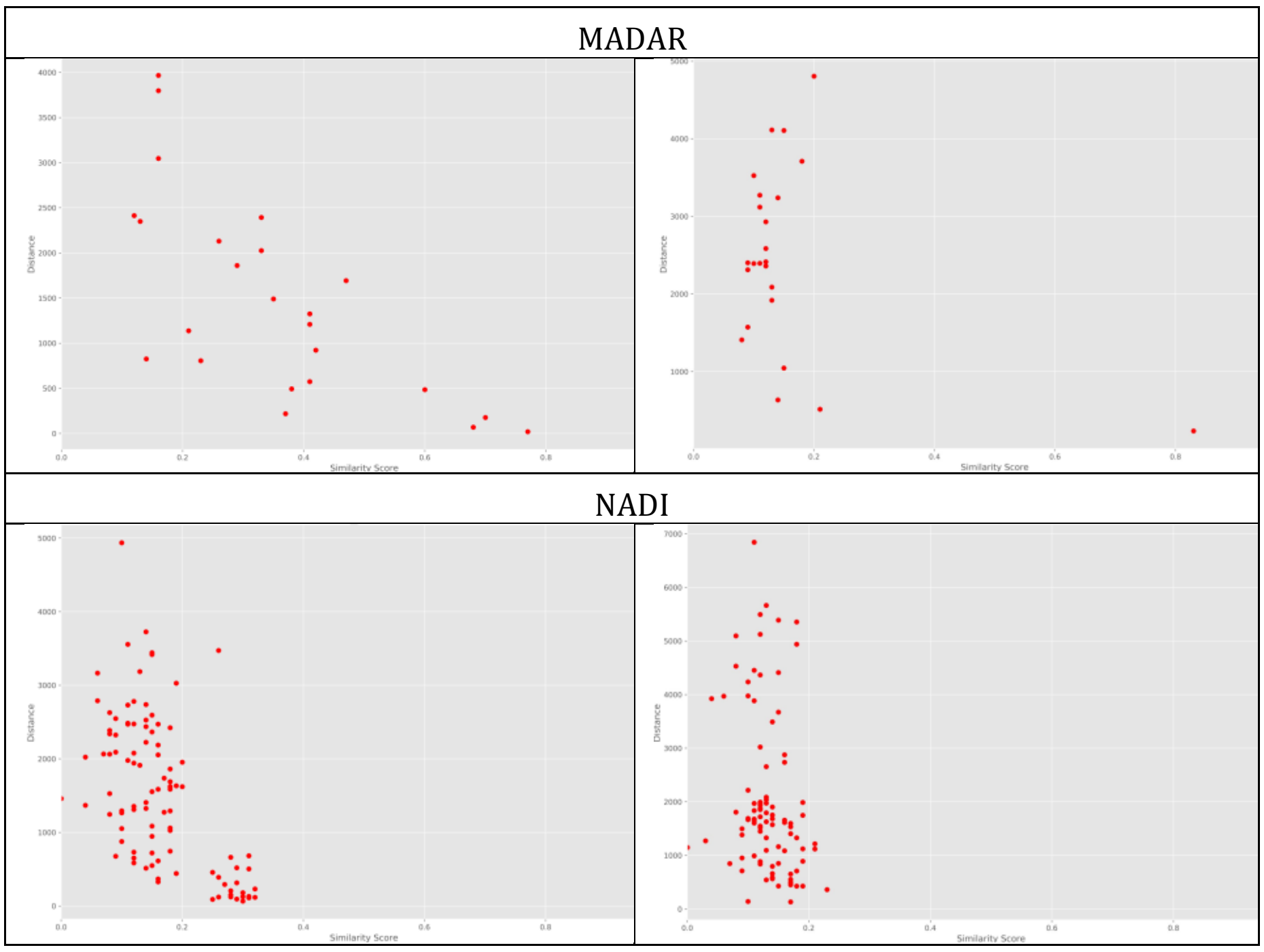

**Figure 3.** Four locations with their similarity values and distances with other locations. The locations are Amman and Tunis for MADAR and Cairo and Bahrain for NADI.

With evidence that there were locations with similarity values higher than the average within the same country, and evidence of several pairs not from the same country with a statistically significant positive correlation, we then wanted to examine whether distance could be correlated with similarity. To examine this, we tested whether a decrease in distance was correlated with an increase in similarity by calculating the distances between pairs as described in Section 4.4. We then added the results to the dataset. Therefore, each pair now has a similarity value, a correlation value, and a distance value. In this section, we will analyse individual cities and examine whether there is a correlation between their similarity values and their distances from other locations. the correlation between the distances for each city, as well as the similarity values for each city with all the other cities or provinces.

For each dataset, we processed the results for each city. These results included the city's similarity and distance values with all the other locations. Subsequently, we applied the Pearson correlation coefficient test to identify whether similarity was correlated with distance for each location. Figure 3 shows a sample of these results. Each sub-figure shows the similarity and distance values (in km) for a selected city. As the Figure shows, there seems to be a correlation between the two variables for some of the cities. For example, the first image for MADAR shows the values for Amman, Jordan. The figure demonstrates a clear correlation for the city, as similarity values increased with the decrease in distance. However, the case was not the same for the second image, which shows Tunis, Tunisia. Most values for the city were under .25 similarity. The only city that exceeded that value was also located in Tunisia. Amman had a correlation of -0.70 and p-value of 0.0001, while the values were -0.36 and 0.07 for Tunis. For NADI, the Figure shows Cairo first, followed by the capital of Bahrain. Cairo had a correlation value of -.57 and p-value that was well below 0.01, while the second location had a correlation of -0.22 and a p-value of 0.02. Finally, we quantified the number of cities in each dataset with a statistically significant correlation between their distance and similarity values. For MADAR, results indicated that 18 of the 25 cities had a SS correlation at the 0.05 level. The results were similar for NADI, with 68 of the 100 provinces meeting the defined criteria. Therefore, the percentages for both were 72% for MADAR and 68% for NADI. These results provided a further indication of the role that distance plays in determining similarity.

## *5.5. Investigating similarity and distance for the entire sets (RQ5)*

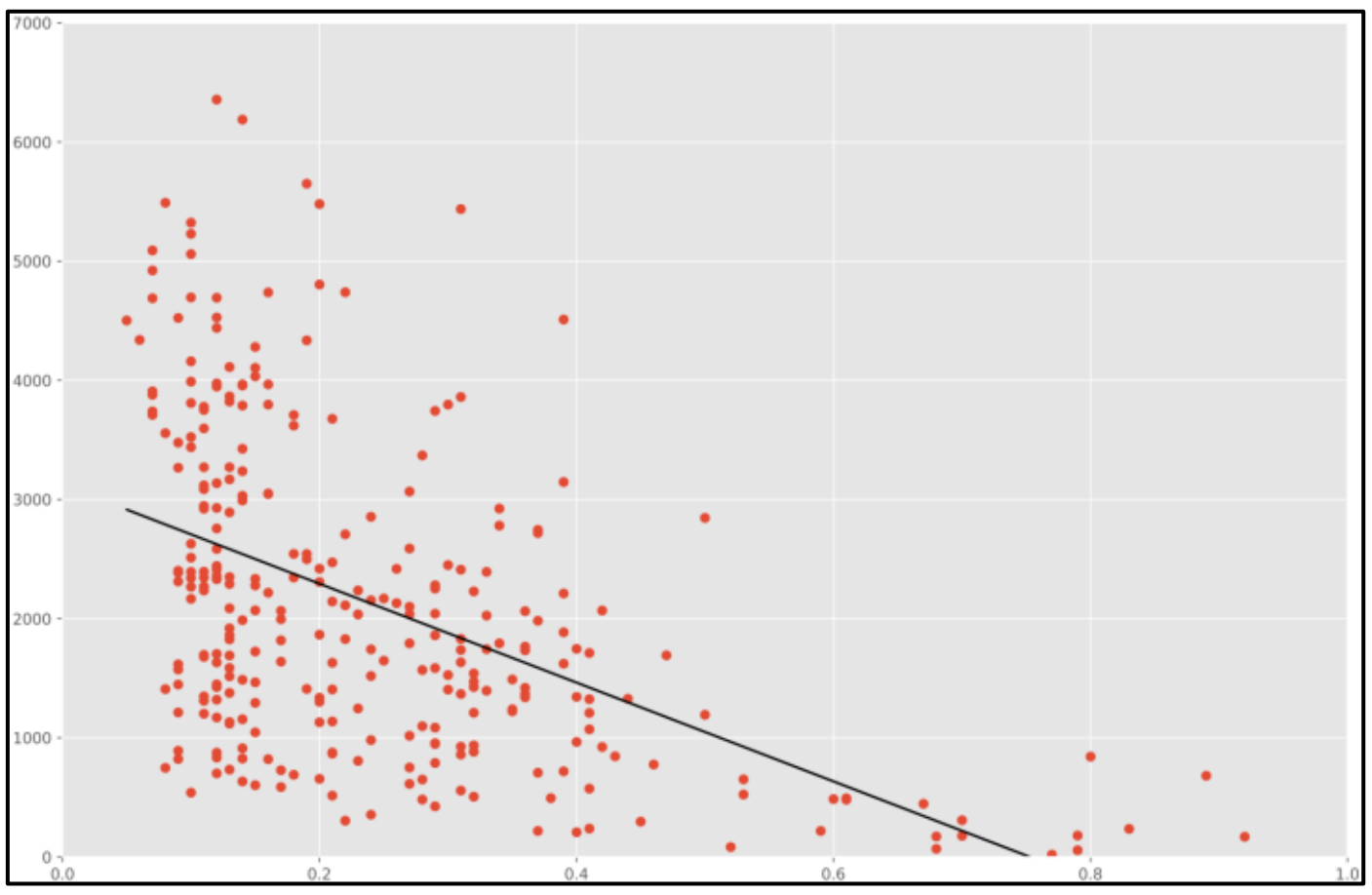

**Figure 4.** Similarity and Distance for all pairs in MADAR

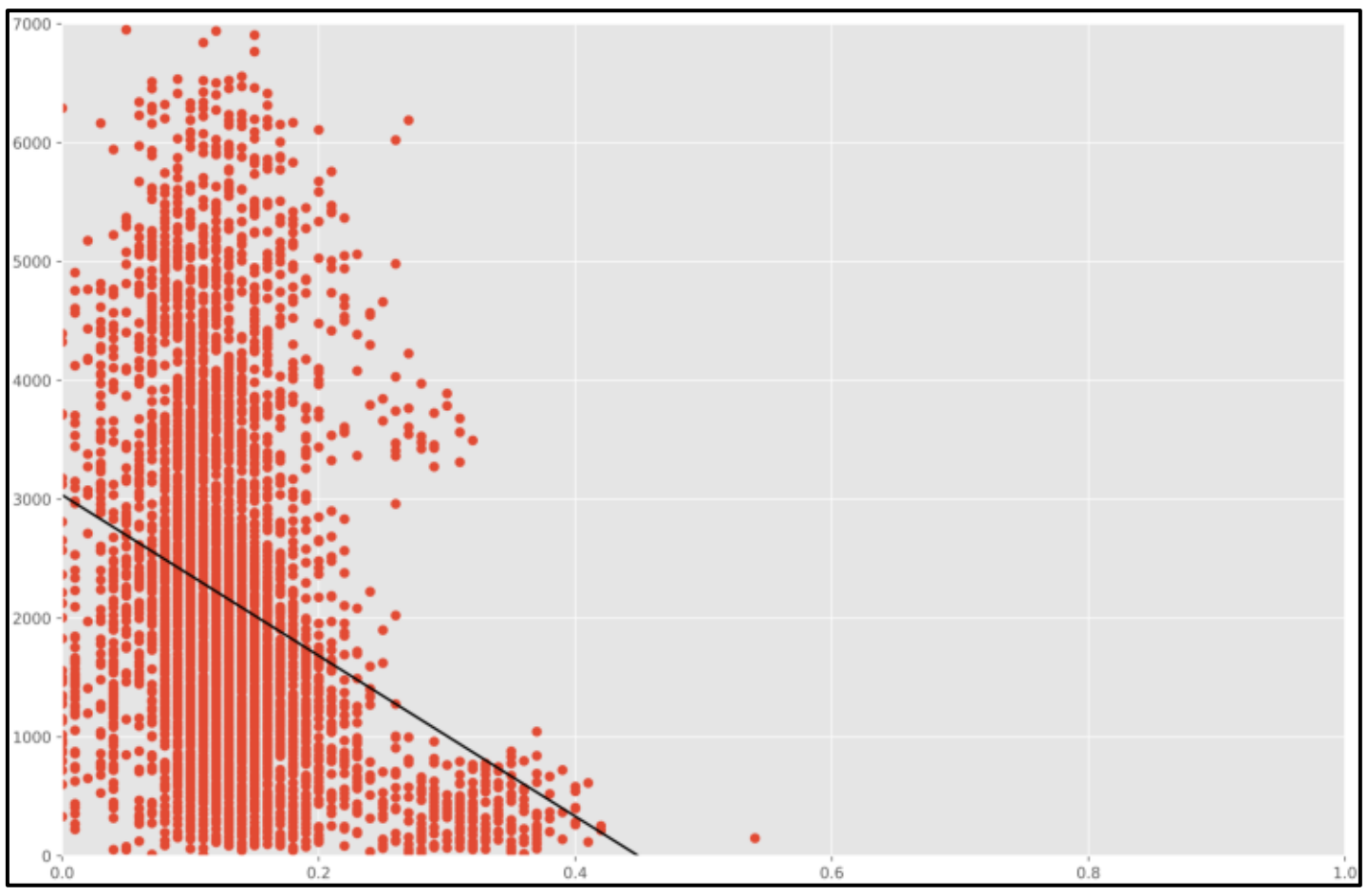

**Figure 5.** Similarity and Distance for all pairs in NADI

For the final group of results, we processed all the pairs with their similarity and distance values. For the 300 pairs in MADAR, the average distance between pairs was 2126.2 km while it was 2142.3 km for the 4950 pairs in NADI. The highest distance between pairs was 6,358 km (Rabat, Morocco & Muscat, Oman) in MADAR and 7,677 km (Nouakchott, Mauritania & Ash Sharqiyah, Oman) in NADI. As for the shortest distance, the shortest distance in MADAR was 20 km (Salt & Amman in Jordan) and 11 km (Ar Rayyan & Doha in Qatar) in NADI.

First, we performed a test to determine whether similarity and distance were correlated for the two datasets. To discover this, we retrieved all the similarity values and all the distance values for pairs in each dataset. Then, we applied the Pearson correlation coefficient test, with these lists as input (i.e., we ran the test twice, once for each dataset). Figure 4 shows a scatterplot containing all the pairs in MADAR and Figure 5 shows the same for NADI. The figures also include a regression line. The Pearson correlation coefficient value was -.495 for MADAR and -.289 for NADI. In both cases, the results were below the 0.01 SL. Thus, results indicate that there is indeed a statistically significant correlation between distance and similarity between Arabic dialects. This demonstrates that the two variables are correlated regardless of whether city pairs were found in the same country. The implications for these findings include perhaps a rethinking of the framing of city-level Arabic dialect identification. The two figures provide a clear illustration of relevant differences between the two datasets. While most points in NADI were below the .2 similarity threshold, many pairs in MADAR seemed to have higher values. Most importantly though, the figures show that there is a correlation between distance and similarity in the two datasets.

## *5.6. Investigating similarity and distance for groups (RQ6)*

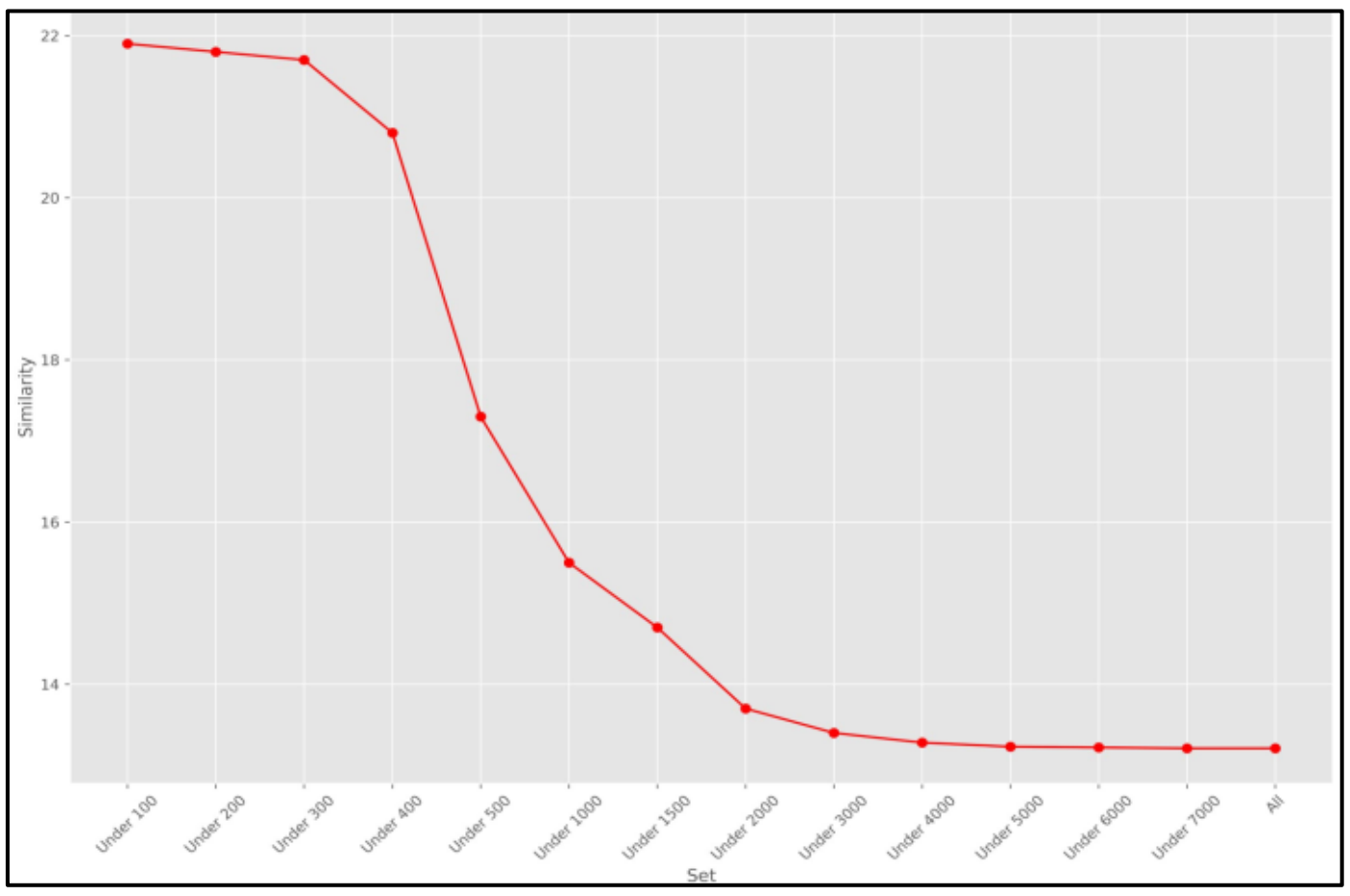

**Figure 6.** The sets in MADAR and their average similarity

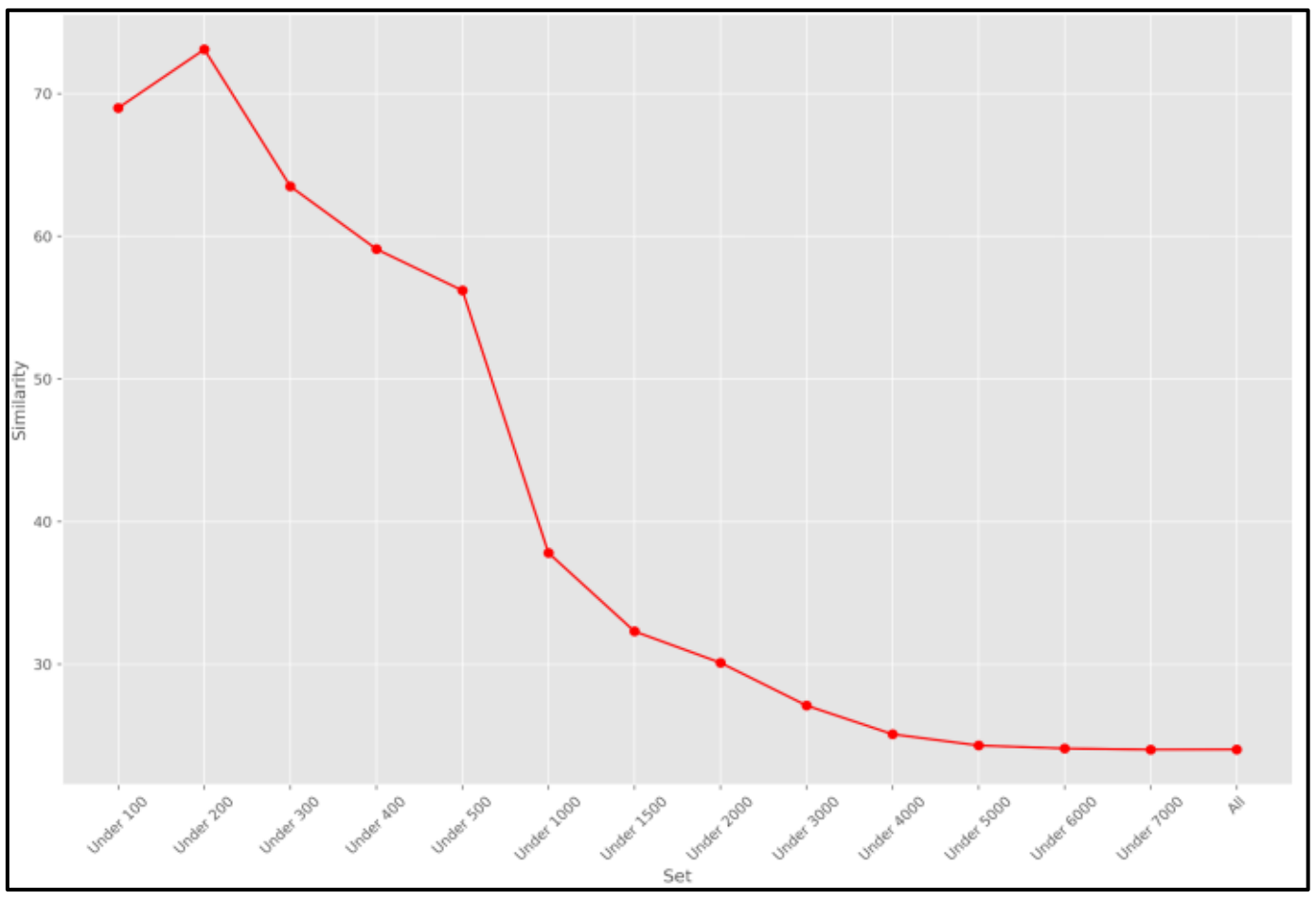

**Figure 7.** The sets in NADI and their average similarity

The final research question is concerned with dividing the pairs in each dataset into groups based on defined ranges of distance. The groups were placed in 13 different sets, shown in Figures 6 and 7. For example, “under 100” included pairs with distance below 100 km while "under 200” included those with distance below 200 km (including the pairs that were under 100 km). The points displayed represent the average similarity values for all the entries in the group. For example, there were 14 pairs with distance values that were less than 300 km. The average similarity for pairs in this set was 63.5%. The exact values for the averages and the number of pairs in each set are listed in Tables 6 and 7. The "All" group that is displayed at the very right of the figures shows the averages for all the pairs. Although the two figures demonstrate how the two datasets vary with regard to their average similarity values, they nevertheless illustrate a similar trend. For both, the first groups show high similarities before expressing a major drop that is followed by a gradual decrease. This decrease is followed by sets where the averages eventually converge. In summary, these findings respond to the sixth research question, demonstrating that there is indeed a trend where similarity values decrease as the distances between pairs increase.

Table 6: The sets in MADAR and their average similarity

| Set | <100 | <200 | <300 | <400 | <500 | <1000 | <1500 | <2000 | <3000 | <4000 | <5000 | <6000 | <7000 |
|---|---|---|---|---|---|---|---|---|---|---|---|---|---|
| Count | 4 | 8 | 14 | 17 | 24 | 69 | 115 | 157 | 226 | 271 | 290 | 298 | 300 |
| Avg Sim | 69% | 73.1% | 63.5% | 59.1% | 56.1% | 37.8% | 32.3% | 30.1% | 27.1% | 25% | 24.3% | 24% | 24% |

Table 7: The sets in NADI and their average similarity

| Set | <100 | <200 | <300 | <400 | <500 | <1000 | <1500 | <2000 | <3000 | <4000 | <5000 | <6000 | <7000 |
|---|---|---|---|---|---|---|---|---|---|---|---|---|---|
| Count | 48 | 155 | 251 | 373 | 509 | 1187 | 1951 | 2659 | 3775 | 4317 | 4677 | 4876 | 4939 |
| Avg Sim | 21.9% | 21.8% | 21.7% | 20.8% | 20% | 17.3% | 15.5% | 14.7% | 13.7% | 13.4% | 13.2% | 13.2% | 13.2% |

## 6. Discussion

In this study, we explored the effects of distance on the similarity between Arabic dialects from several cities and provinces. The key findings of this work are 1) a demonstration of the correlation between distance and similarity for individual locations and for the entire sets, 2) proof that it is possible for cities that are not located in the same country to be correlated and have similarity values that are higher than the averages for cities located in the same country, and 3) an explanation of a decreasing trend as similarity values between pairs decrease with the increase in distance. Put differently, the further two locations are from each other, the more likely they are to have less similarity in their dialects. It is, however, still important to note that this work has several limitations. First, the work relies on two datasets that have their own limitations, which may have affected the results. For example, the authors of NADI acknowledge that the dataset contains a small percentage of texts that are not written in Arabic [5] and that "province-level classification is somewhat related to geolocation prediction exploiting Twitter data" [6]. As for MADAR, the dataset includes texts from only the travel domain. Moreover, the dataset was "commissioned and not naturally occurring" [6]. Second, the method we used to measure distances between locations is sensitive to the coordinates we used for locations. While this may not be an issue for small cities, it might prove to be problematic for larger provinces where the longitude and latitude pairs we used in the study may not correlate to the origin point of most of the texts. Finally, we only used a single method to measure the similarity between the texts of the dialects.

Our work has several implications. Most importantly, we believe our findings may help better understand how to frame the problem of Arabic dialects classification. Since the task is new and still in its infancy, changing how it is addressed in the literature is possible. Therefore, novel insights regarding specific characteristics of the problem may be useful in rethinking how to develop and evaluate solutions for the task. Based on our findings, we discovered that geographical proximity affects dialect similarity. Therefore, incorporating some aspects of geographical proximity when studying Arabic dialects may help improve methods that target this problem. Moreover, research in this area may benefit from scholarly work on the prediction of the locations of users on various social media platforms [42,43]. In Abdul-Mageed et. al. [28], the authors reviewed some of the relevant work in that domain. Furthermore, they argued that the tasks of identifying the Arabic dialects in Twitter messages and the prediction of the locations of users are actually quite different, despite being related. They indicated that in the former case, the objective is identifying the dialect regardless of the actual physical locations of the users, whereas in the latter, the user's specific location is paramount. Moreover, the task is concerned with processing individual tweets or messages. Although there is room for debate regarding the differences between the two tasks, city-level Arabic dialect identification can still use some of the established evaluation criteria, for example. Since our findings show the impact of the geographical proximity on dialect similarity, it might be useful to consider using the concept of "tolerance distances" in location prediction [44,45]. "Tolerance distance" is employed in order to measure the performance of models that predict locations by specifying ranges of distances (e.g., under 100 km, under 200 km. etc.). Therefore, when a model does not predict the correct location but instead predicts the location of a neighbouring city, the evaluations will capture this difference and will not simply tag the prediction as inaccurate. We believe that our work shows that there are other ways of looking into this research challenge and thus, these other ways should be explored and investigated.

## 7. Conclusion

In this paper, we studied effects of geographical proximity on similarity between Arabic dialects at the city and province levels. Our findings indicate that there is a correlation between the similarity and distance for individual locations as well as for entire sets. Our work provides a new perspective into how to investigate the research challenge of Arabic dialect identification. There are several avenues for future work. First, other similar corpora can also be explored. We can apply the same methodological steps on these new corpora and investigate whether new findings confirm the results discussed in this work. This may also help identify additional relevant observations on which future research can build. Second, incorporating distance in Arabic dialect identification provides another potential avenue for future work. For example, the objective would be to evaluate models not only on their prediction, but also on how geographically close the wrong prediction is compared to the correct one. Finally, the issue of sub dialects in selected cities is another potential area for future work. For example, in larger cities, it is possible that there is a main dialect commonly used in the area and also several sub-dialects used within communities in the city.

## References

[1] M. Abdul-mageed, Modeling Arabic subjectivity and sentiment in lexical space, Inf. Process. Manag. 56 (2019) 291–307. https://doi.org/10.1016/j.ipm.2017.07.004.

[2] I.A. Farha, W. Magdy, A comparative study of effective approaches for Arabic sentiment analysis, Inf. Process. Manag. 58 (2021) 102438. https://doi.org/10.1016/j.ipm.2020.102438.

[3] S. Harrat, K. Meftouh, K. Smaili, Machine translation for Arabic dialects (survey), Inf. Process. Manag. 56 (2019) 262–273. https://doi.org/10.1016/j.ipm.2017.08.003.

[4] K. Darwish, N. Habash, M. Abbas, H. Al-Khalifa, H.T. Al-Natsheh, H. Bouamor, K. Bouzoubaa, V. Cavalli-Sforza, S.R. El-Beltagy, W. El-Hajj, M. Jarrar, H. Mubarak, A Panoramic Survey of Natural Language Processing in

the Arab World, Commun. ACM. 64 (2021) 72–81. https://doi.org/10.1145/3447735.

[5] M. Abdul-Mageed, C. Zhang, H. Bouamor, N. Habash, NADI 2020: The First Nuanced Arabic Dialect Identification Shared Task, in: Proc. Fifth Arab. Nat. Lang. Process. Work., Association for Computational Linguistics, Barcelona, Spain (Online), 2020: pp. 97–110. https://www.aclweb.org/anthology/2020.wanlp-1.9.

[6] M. Abdul-Mageed, C. Zhang, A. Elmadany, H. Bouamor, N. Habash, NADI 2021: The Second Nuanced Arabic Dialect Identification Shared Task, in: Proc. Sixth Arab. Nat. Lang. Process. Work. (WANLP 2021), 2021.

[7] A. Alshutayri, E. Atwell, Exploring Twitter as a source of an Arabic dialect corpus, Int. J. Comput. Linguist. 8 (2017) 37–44.

[8] F. Sadat, F. Kazemi, A. Farzindar, Automatic Identification of Arabic Dialects in Social Media, in: Proc. First Int. Work. Soc. Media Retr. Anal., Association for Computing Machinery, New York, NY, USA, 2014: pp. 35–40. https://doi.org/10.1145/2632188.2632207.

[9] Y. Khalifa, A. Elnagar, Colloquial Arabic Tweets: Collection, Automatic Annotation, and Classification, in: 2020 Int. Conf. Asian Lang. Process., 2020: pp. 163–168. https://doi.org/10.1109/IALP51396.2020.9310507.

[10] 639 Identifier Documentation: ARA, SIL Int. (n.d.). https://iso639-3.sil.org/code/ara.

[11] A. Abdelali, H. Mubarak, Y. Samih, S. Hassan, K. Darwish, QADI: Arabic Dialect Identification in the Wild, in: Proc. Sixth Arab. Nat. Lang. Process. Work., Association for Computational Linguistics, Kyiv, Ukraine (Virtual), 2021: pp. 1–10. https://www.aclweb.org/anthology/2021.wanlp-1.1.

[12] S. Al-hanouf, S. Alshalan, R. Alshalan, N. Alrumayyan, S. Al-manea, N. Al-twairesh, R. Al-matham, N. Madi, N. Almugren, S. Bawazeer, N. Al-mutlaq, N. Almanea, W. Bin Huwaymil, S. Alshalan, R. Alshalan, N. Alrumayyan, S. Al-manea, R. Alotaibi, S. Al-senaydi, A. Alfutamani, S. Bawazeer, N. Al-

mutlaq, N. Almanea, W. Bin Huwaymil, D. Alqusair, R. Alotaibi, S. Al-Senaydi, A. Alfutamani, SUAR: Towards Building a Corpus for the Saudi Dialect, Procedia Comput. Sci. 142 (2018) 72–82. https://doi.org/10.1016/j.procs.2018.10.462.

[13] H. Bouamor, N. Habash, M. Salameh, W. Zaghouani, O. Rambow, D. Abdulrahim, O. Obeid, S. Khalifa, F. Eryani, A. Erdmann, K. Oflazer, The MADAR Arabic Dialect Corpus and Lexicon, in: LREC, 2018.

[14] K. Abu Kwaik, M. Saad, S. Chatzikyriakidis, S. Dobnik, Shami: A Corpus of Levantine Arabic Dialects, in: Proc. Elev. Int. Conf. Lang. Resour. Eval. 2018, European Language Resources Association (ELRA), Miyazaki, Japan, 2018.

[15] R. Ziedan, M. Micheal, A. Alsammak, M. Mursi, A. Elmaghraby, A unified approach for arabic language dialect detection, in: 29th Int. Conf. Comput. Appl. Ind. Eng. (CAINE 2016), Denver, USA, 2016.

[16] A. Alshutayri, E. Atwell, Arabic dialects annotation using an online game, in: 2018 2nd Int. Conf. Nat. Lang. Speech Process., 2018: pp. 1–5. https://doi.org/10.1109/ICNLSP.2018.8374371.

[17] K. Almeman, Automatically Building VoIP Speech Parallel Corpora for Arabic Dialects, ACM Trans. Asian Low-Resour. Lang. Inf. Process. 17 (2017). https://doi.org/10.1145/3132708.

[18] R. Zarnoufi, H. Jaafar, M. Abik, Machine Normalization: Bringing Social Media Text from Non-Standard to Standard Form, ACM Trans. Asian Low-Resour. Lang. Inf. Process. 19 (2020). https://doi.org/10.1145/3378414.

[19] R. Tachicart, K. Bouzoubaa, S.L. Aouragh, H. Jaafa, Automatic identification of Moroccan colloquial Arabic, in: Int. Conf. Arab. Lang. Process., 2017: pp. 201–214.

[20] M. Lichouri, M. Abbas, A.A. Freihat, D.E.H. Megtouf, Word-Level vs Sentence-Level Language Identification: Application to Algerian and Arabic Dialects, Procedia Comput. Sci. 142 (2018) 246–253. https://doi.org/https://doi.org/10.1016/j.procs.2018.10.484.

[21] A. Masmoudi, S. Mdhaffar, R. Sellami, L.H. Belguith, Automatic Diacritics Restoration for Tunisian Dialect, ACM Trans. Asian Low-Resour. Lang. Inf. Process. 18 (2019). https://doi.org/10.1145/3297278.

[22] A. Alnawas, N. Arici, Sentiment Analysis of Iraqi Arabic Dialect on Facebook Based on Distributed Representations of Documents, ACM Trans. Asian Low-Resour. Lang. Inf. Process. 18 (2019). https://doi.org/10.1145/3278605.

[23] M. Al-ayyoub, A. Allah, Y. Jararweh, M.N. Al-kabi, M. Sa, A comprehensive survey of arabic sentiment analysis, Inf. Process. Manag. 56 (2019) 320–342. https://doi.org/10.1016/j.ipm.2018.07.006.

[24] W. Farhan, B. Talafha, A. Abuammar, R. Jaikat, M. Al-ayyoub, A.B. Tarakji, A. Toma, R. Samsung, Unsupervised dialectal neural machine translation, Inf. Process. Manag. 57 (2020) 102181. https://doi.org/10.1016/j.ipm.2019.102181.

[25] A. Soufan, Deep Learning for Sentiment Analysis of Arabic Text, in: Proc. Arab. 6th Annu. Int. Conf. Res. Track, Association for Computing Machinery, New York, NY, USA, 2019. https://doi.org/10.1145/3333165.3333185.

[26] Q. Zhang, J.H.L. Hansen, Language/Dialect Recognition Based on Unsupervised Deep Learning, IEEE/ACM Trans. Audio, Speech and Lang. Proc. 26 (2018) 873–882. https://doi.org/10.1109/TASLP.2018.2797420.

[27] M. Alali, N. Mohd Sharef, M.A. Azmi Murad, H. Hamdan, N.A. Husin, Narrow Convolutional Neural Network for Arabic Dialects Polarity Classification, IEEE Access. 7 (2019) 96272–96283. https://doi.org/10.1109/ACCESS.2019.2929208.

[28] M. Abdul-Mageed, C. Zhang, A. Elmadany, L. Ungar, Toward micro-dialect identification in diaglossic and code-switched environments, Proc. 2020 Conf. Empir. Methods Nat. Lang. Process. (2020).

[29] O. Obeid, M. Salameh, H. Bouamor, N. Habash, ADIDA: Automatic Dialect Identification for Arabic, in: Proc. 2019 Conf. North {A}merican Chapter Assoc. Comput. Linguist., Association for Computational Linguistics,

Minneapolis, Minnesota, 2019: pp. 6–11. https://doi.org/10.18653/v1/N19-4002.

[30] C. Zhang, M. Abdul-Mageed, No Army, No Navy: BERT Semi-Supervised Learning of Arabic Dialects, in: Proc. Fourth Arab. Nat. Lang. Process. Work., Association for Computational Linguistics, Florence, Italy, 2019: pp. 279–284. https://doi.org/10.18653/v1/W19-4637.

[31] P. Mishra, V. Mujadia, Arabic dialect identification for travel and twitter text, in: Proc. Fourth Arab. Nat. Lang. Process. Work., 2019: pp. 234–238.

[32] A. Ragab, H. Seelawi, M. Samir, A. Mattar, H. Al-Bataineh, M. Zaghloul, A. Mustafa, B. Talafha, A.A. Freihat, H. Al-Natsheh, Mawdoo3 ai at madar shared task: Arabic fine-grained dialect identification with ensemble learning, in: Proc. Fourth Arab. Nat. Lang. Process. Work., 2019: pp. 244–248.

[33] B. Talafha, M. Ali, M.E. Za'ter, H. Seelawi, I. Tuffaha, M. Samir, W. Farhan, H. Al-Natsheh, Multi-dialect Arabic BERT for Country-level Dialect Identification, in: Proc. Fifth Arab. Nat. Lang. Process. Work., Association for Computational Linguistics, Barcelona, Spain (Online), 2020: pp. 111–118. https://www.aclweb.org/anthology/2020.wanlp-1.10.

[34] A. El Mekki, A. Alami, H. Alami, A. Khoumsi, I. Berrada, Weighted combination of BERT and N-GRAM features for Nuanced Arabic Dialect Identification, in: Proc. Fifth Arab. Nat. Lang. Process. Work., Association for Computational Linguistics, Barcelona, Spain (Online), 2020: pp. 268–274.

[35] N. AlShenaifi, A. Azmi, Faheem at NADI shared task: Identifying the dialect of Arabic tweet, in: Proc. Fifth Arab. Nat. Lang. Process. Work., Association for Computational Linguistics, Barcelona, Spain (Online), 2020: pp. 282–287. https://www.aclweb.org/anthology/2020.wanlp-1.29.

[36] K.A. Kwaik, M. Saad, S. Chatzikyriakidis, S. Dobnika, A Lexical Distance Study of Arabic Dialects, Procedia Comput. Sci. 142 (2018) 2–13. https://doi.org/https://doi.org/10.1016/j.procs.2018.10.456.

[37] H. Bouamor, N. Habash, K. Oflazer, A Multidialectal Parallel Corpus of

Arabic., in: LREC, 2014: pp. 1240–1245.

[38] S. Bird, E. Klein, E. Loper, Natural Language Processing with Python, 1st ed., O'Reilly Media, Inc., 2009.

[39] F. Pedregosa, G. Varoquaux, A. Gramfort, V. Michel, B. Thirion, O. Grisel, M. Blondel, P. Prettenhofer, R. Weiss, V. Dubourg, Scikit-learn: Machine learning in Python, J. Mach. Learn. Res. 12 (2011) 2825–2830.

[40] P. Virtanen, R. Gommers, T.E. Oliphant, M. Haberland, T. Reddy, D. Cournapeau, E. Burovski, P. Peterson, W. Weckesser, J. Bright, S.J. van der Walt, M. Brett, J. Wilson, K.J. Millman, N. Mayorov, A.R.J. Nelson, E. Jones, R. Kern, E. Larson, C.J. Carey, I. Polat, Y. Feng, E.W. Moore, J. VanderPlas, D. Laxalde, J. Perktold, R. Cimrman, I. Henriksen, E.A. Quintero, C.R. Harris, A.M. Archibald, A.H. Ribeiro, F. Pedregosa, P. van Mulbregt, SciPy 1.0: fundamental algorithms for scientific computing in Python, Nat. Methods. 17 (2020) 261–272. https://doi.org/10.1038/s41592-019-0686-2.

[41] Haversine: Calculate the distance (in various units) between two points on Earth using their latitude and longitude, (2020). https://github.com/mapado/haversine.

[42] X. Luo, Y. Qiao, C. Li, J. Ma, Y. Liu, An overview of microblog user geolocation methods, Inf. Process. Manag. 57 (2020) 102375. https://doi.org/https://doi.org/10.1016/j.ipm.2020.102375.

[43] J.D.G. Paule, Y. Sun, Y. Moshfeghi, On fine-grained geolocalisation of tweets and real-time traffic incident detection, Inf. Process. Manag. 56 (2019) 1119–1132. https://doi.org/https://doi.org/10.1016/j.ipm.2018.03.011.

[44] O. Ozdikis, H. Ramampiaro, K. Nørvåg, Locality-adapted kernel densities of term co-occurrences for location prediction of tweets, Inf. Process. Manag. 56 (2019) 1280–1299. https://doi.org/https://doi.org/10.1016/j.ipm.2019.02.013.

[45] P. Zola, C. Ragno, P. Cortez, A Google Trends spatial clustering approach for a worldwide Twitter user geolocation, Inf. Process. Manag. 57 (2020) 102312. https://doi.org/https://doi.org/10.1016/j.ipm.2020.102312.